\pdfoutput=1
\documentclass[letterpaper, 10pt, conference]{ieeeconf}

\IEEEoverridecommandlockouts
\usepackage[T1]{fontenc}        % Type 1 font encoding (avoids Type 3 in PDF)
\usepackage{mathptmx}           % Times New Roman for text + math
\DeclareMathAlphabet{\mathcal}{OMS}{cmsy}{m}{n}
\usepackage{bm}                 % Solving the problem of mathptmx not being able to bold mathematical symbols
\usepackage{graphicx}
\usepackage{amsmath}
\usepackage{amssymb}
\usepackage{cite}
\usepackage{url}
\usepackage{booktabs}
\usepackage{multirow}
\usepackage{pifont}
\usepackage{array}
\usepackage{hyperref}
\usepackage{stfloats}
\usepackage{dsfont}

\newcount\irospagelimit \irospagelimit=8
\AtEndDocument{%
  \ifnum\value{page}>\irospagelimit
    \GenericWarning{}{IROS FORMAT WARNING: Paper is \the\value{page} pages,^^J%
      exceeding the IROS 2026 limit of \the\irospagelimit\space pages!^^J%
      You MUST reduce to \the\irospagelimit\space pages or fewer before submission.}%
  \fi
}

\usepackage{censor}
\StopCensoring  % Comment out this line for anonymous submission

\usepackage{microtype}          % Subliminal refinements: protrusion + expansion

\let\oldthebibliography\thebibliography
\renewcommand{\thebibliography}[1]{%
  \oldthebibliography{#1}%
  \setlength{\itemsep}{0pt}%
  \setlength{\parsep}{0pt}%
  \setlength{\parskip}{0pt}%
}

\title{\LARGE \bf
Omnidirectional Humanoid Locomotion on Stairs via Unsafe Stepping Penalty and Sparse LiDAR Elevation Mapping
}

\author{Yuzhi Jiang$^{1}$, Yujun Liang$^{1}$, Junhao Li$^{1}$, Han Ding$^{2,3}$, Lijun Zhu$^{1,2,\dagger}$% <-this % stops a space
\thanks{This work was supported in part by the Fundamental and Interdisciplinary Disciplines Breakthrough Plan of the Ministry of Education of China under Grant JYB2025XDXM208, and in part by the National Natural Science Foundation of China under Grant U25A6013.}%
\thanks{$^{\dagger}$Corresponding author.}%
\thanks{$^{1}$School of Artificial Intelligence and Automation, Huazhong University of Science and Technology, Wuhan 430074, China. {\tt\{yzjiang, andrew, junhaoli, ljzhu\}@hust.edu.cn}}%
\thanks{$^{2}$State Key Laboratory of Intelligent Manufacturing Equipment and Technology, Huazhong University of Science and Technology, Wuhan 430074, China. {\tt dinghan@mail.hust.edu.cn}}%
\thanks{$^{3}$School of Mechanical Science and Engineering, Huazhong University of Science and Technology, Wuhan 430074, China.}%
}

\begin{document}

\maketitle
\thispagestyle{empty}

%%%%%%%%%%%%%%%%%%%%%%%%%%%%%%%%%%%%%%%%%%%%%%%%%%% Title above %%%%%%%%%%%%%%%%%%%%%%%%%%%%%%%%%%%%%%%%%%%%%%%%%%%

\begin{abstract}

Humanoid robots, characterized by numerous degrees of freedom and a high center of gravity, are inherently unstable. Safe omnidirectional locomotion on stairs requires both omnidirectional terrain perception and reliable foothold selection. Existing methods often rely on forward-facing depth cameras, which create blind zones that restrict omnidirectional mobility. Furthermore, most existing safe-foothold rewards rely on sparse post-contact signals, which slow convergence. Building on these rewards, this paper introduces a single-stage training framework with a dense unsafe stepping penalty that provides continuous pre-contact feedback as the foot approaches a hazardous placement. We further build a rolling point-cloud mapping system with spatiotemporal confidence decay and a self-protection zone mechanism for temporally consistent local maps. These maps are further refined by an Edge-Guided Asymmetric U-Net (EGAU), which mitigates reconstruction distortion caused by sparse LiDAR returns on stair risers. Simulation and real-robot experiments show that the proposed method substantially reduces the probability of unsafe foot placements on stair terrains and completes a continuous long-distance walking test on complex outdoor terrains, demonstrating reliable sim-to-real transfer and long-term stability.

\end{abstract}

%%%%%%%%%%%%%%%%%%%%%%%%%%%%%%%%%%%%%%%%%%%%%%%%%%% Abstract above %%%%%%%%%%%%%%%%%%%%%%%%%%%%%%%%%%%%%%%%%%%%%%%%%%%

\section{INTRODUCTION}

Humanoid robots are attractive for domestic service, post-disaster response, and industrial inspection because their human-scale, bipedal form lets them move through environments built for people, such as stairs and curbs. Yet their high center of mass and intermittent ground support make them prone to losing balance on stairs, where each step demands precise foot placement \cite{LBL}.

\begin{figure}[!t]
\centering
\includegraphics[width=0.95\columnwidth]{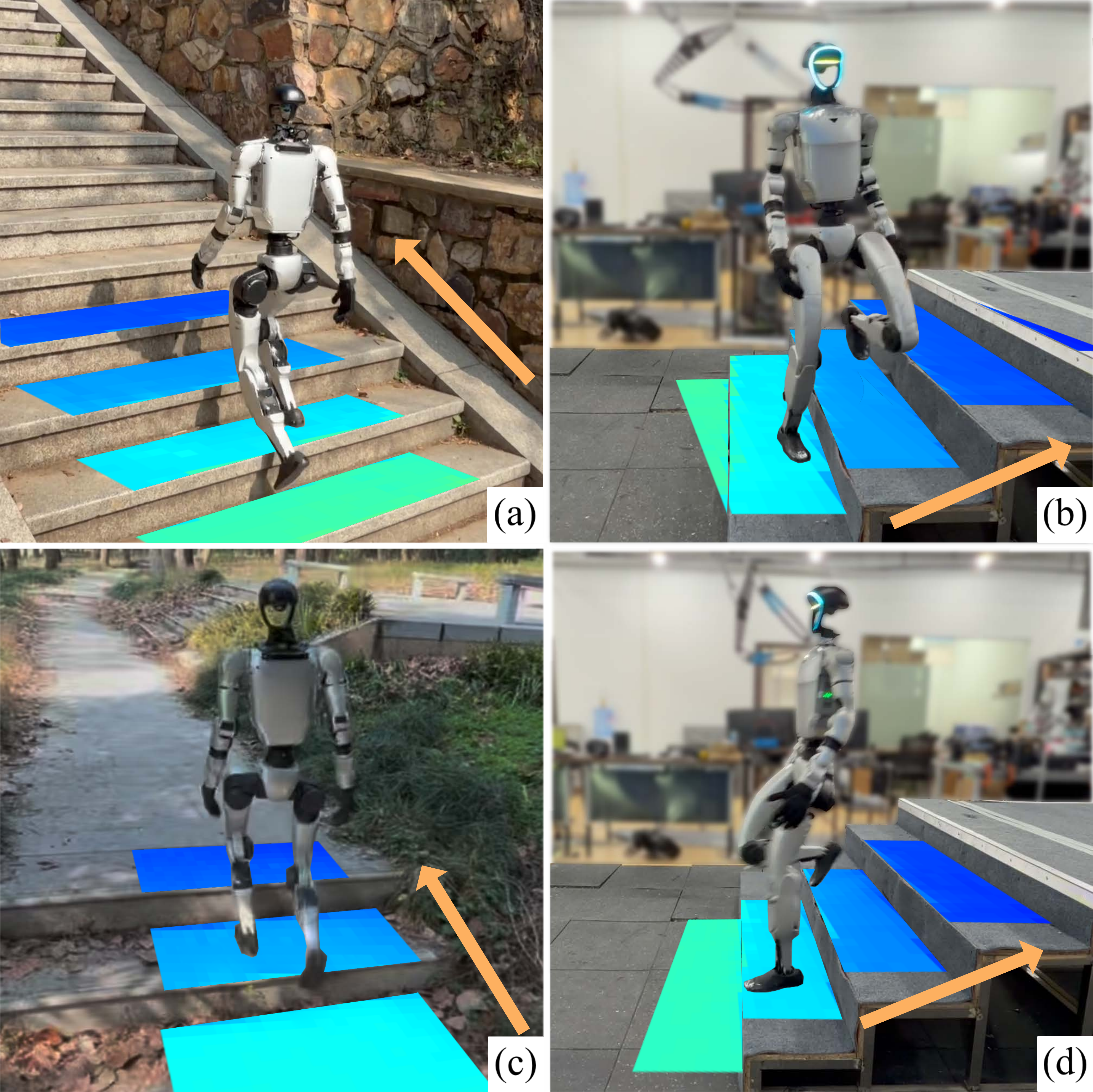}
\caption{Overview of the proposed omnidirectional perceptive locomotion framework deployed on the Unitree G1 humanoid robot. The robot safely performs omnidirectional stair traversal in both indoor and outdoor settings, including forward ascent (a), (c), lateral ascent (b), and backward ascent (d). Orange arrows indicate the direction of motion, and the teal overlay on the stairs illustrates the real-time reconstructed elevation map.}
\label{fig:teaser}
\end{figure}

Model-based control explicitly encodes dynamics and constraints, but high-DoF stair locomotion is sensitive to contact timing, terrain geometry, and impacts that are difficult to model under sparse perception and frequent contact switches. Learning-based policies can instead optimize closed-loop whole-body behavior over randomized terrains and contacts. Yet proprioception-only policies remain insufficient for tall stairs \cite{robot_parkour_learning, attention_map}, making external perception necessary.

Depth cameras have enabled impressive vision-based locomotion, and recent single-stage frameworks such as PIE~\cite{pie} have substantially narrowed the depth sim-to-real gap. Nevertheless, head-mounted frontal depth inherently leaves lateral and rear footholds poorly observed during sideways or backward stair traversal. For omnidirectional foothold coverage, we adopt a LiDAR-based omnidirectional elevation map.

Guiding safe foothold selection requires carefully designed rewards. One line tracks footholds from a model-based planner~\cite{dtc}; another uses sparse foothold rewards that fire only after a collision or edge contact~\cite{beamdojo,humanoid_parkour}. The former adds planning overhead, while sparse post-contact feedback gives little gradient for precise placement. We instead adopt a dense, pre-contact reward-shaping term that needs no reference trajectory.

To address these challenges, we propose a single-stage omnidirectional perceptive locomotion framework: a dense unsafe stepping penalty (foot-collision and edge-contact terms) gives continuous feedback before a hazardous footstep, while a LiDAR-based robot-centric elevation map provides omnidirectional coverage for confident lateral and backward motion.

In summary, the main contributions of this paper include:
\begin{itemize}
    \item Building on prior foothold-safety rewards, a dense unsafe stepping penalty is designed that supplies continuous pre-contact feedback for both foot-collision and edge-stepping, improving the safe stepping rate and learning efficiency for omnidirectional stair traversal without a model-based reference.
    \item A perceptive architecture is developed that integrates point-cloud-based rolling mapping with an Edge-Guided Asymmetric U-Net (EGAU), effectively addressing the feature reconstruction distortion caused by extremely sparse onboard point clouds and physical blind zones.
    \item Extensive validation in both simulation and real-world demonstrates strong performance in omnidirectional stair traversal and long-distance outdoor walking, confirming notable safety and robust sim-to-real transfer capability.
\end{itemize}
%%%%%%%%%%%%%%%%%%%%%%%%%%%%%%%%%%%%%%%%%%%%%%%%%%% Introduction above %%%%%%%%%%%%%%%%%%%%%%%%%%%%%%%%%%%%%%%%%%%%%%%%%%%

\section{RELATED WORK}

Deep reinforcement learning is now the dominant paradigm for legged locomotion on complex terrain. Early proprioceptive approaches walk stably on flat ground but perform poorly on discontinuous terrain such as stairs, where external perception becomes essential. Many studies use forward-facing depth cameras to obtain depth maps of the frontal terrain \cite{visual_locomotion, cross_modal, nvm, terrain_recon, bipedal_vision}. To ease learning from high-dimensional observations, teacher-student architectures are common \cite{robot_parkour_learning, egocentric_vision, extreme_parkour, soloparkour}, but two-stage distillation can incur information loss. To avoid this, single-stage implicit-explicit frameworks such as PIE~\cite{pie} learn perception and control jointly and show robust illumination- and depth-related sim-to-real transfer. We adopt LiDAR for its omnidirectional field of view, which covers the lateral and backward footholds that forward-only depth cannot observe.

To provide richer perceptive information, numerous studies feed elevation maps into the policy \cite{beamdojo,pim_paper,perceptive_humanoid,robust_perceptive,anymal_parkour}. Classical probabilistic mapping, as represented by Fankhauser et al.~\cite{eth_elevation}, suffers from stale observations and step-edge melting when confronted with the extremely sparse and chronically occluded point clouds produced by onboard LiDAR. Subsequent research introduces learning-based perception modules such as hybrid internal models \cite{pim_paper} or VAE-based denoising \cite{robust_perceptive} to improve robustness. A concurrent effort, MARG~\cite{marg}, learns a terrain-map generation model that produces drift-reduced robot-centric elevation maps from a single LiDAR. As a leading recent work in blind-zone reconstruction, Song et al. \cite{song2025gait} utilize a standard multi-task U-Net. However, the late-branching design lacks explicit geometric priors during progressive decoding and tends to lose critical step-edge features when filling sparse holes. In contrast, our rolling point-cloud map with spatiotemporal confidence decay and a protection-zone mechanism preserves historical observations in blind regions, while EGAU injects geometric boundary priors to block cross-edge smooth interpolation, producing high-quality omnidirectional elevation maps with accurate step-edge features.

Avoiding unsafe foot placements is central to traversing complex terrain, and several complementary mechanisms have been explored. Jenelten et al.~\cite{dtc} track footholds generated by a model-based planner, achieving precise placement on sparse terrain. He et al.~\cite{attention_map} use attention-based map encoding explicitly designed for generalized foothold reasoning across terrains, an approach further strengthened by subsequent attention-based encoders. Wang et al.~\cite{beamdojo} combine a sparse foothold-support reward with a double-critic scheme, Zhuang et al.~\cite{humanoid_parkour} penalize deviation from ideal foothold centers, and MARG~\cite{marg} employs feet-center, feet-stumble, and feet-air-time rewards. Building on this line, our dense unsafe stepping penalty delivers continuous pre-contact feedback via foot-collision and edge terms that steer the foot before a hazardous step. Needing no reference trajectory or dual-critic scheme, it applies uniformly across omnidirectional stair traversal, improving safety and learning efficiency.

%%%%%%%%%%%%%%%%%%%%%%%%%%%%%%%%%%%%%%%%%%%%%%%%% Related works above %%%%%%%%%%%%%%%%%%%%%%%%%%%%%%%%%%%%%%%%%%%%%%%%%

\section{METHODS}

\begin{figure*}[!t]
    \centering
    \includegraphics[width=1.0\textwidth]{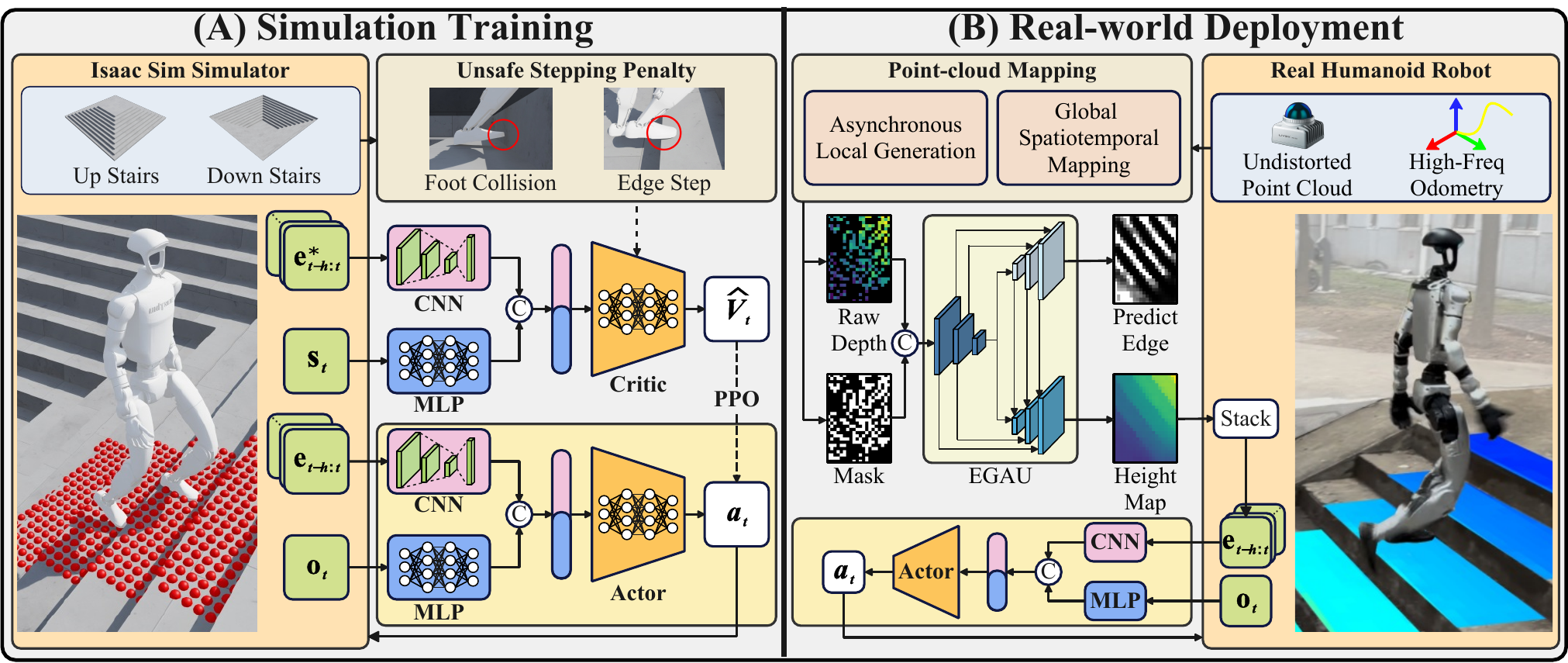}
    \caption{Overview of the proposed framework. (A) Simulation training: an MLP and a CNN extract proprioceptive and perceptive features, respectively, which are then concatenated. Meanwhile, an unsafe foot penalty function guides the policy to learn safe stair-traversal gaits. (B) Real-world deployment: the trained policy network is directly transferred to the real robot, with a point-cloud mapping and edge-guided elevation map module providing stable and accurate elevation map inputs.}
    \label{fig:pipeline}
\end{figure*}

Fig.~\ref{fig:pipeline} illustrates the overall pipeline of the proposed framework, which consists of a simulation training stage and a real-world deployment stage. The following subsections describe each component in detail.

\subsection{Simulation Training}

Owing to the limitations of onboard sensors, the robot cannot access the complete ground-truth state of itself or the surrounding environment. We therefore formulate the reinforcement learning problem as a Partially Observable Markov Decision Process (POMDP), described by the tuple $\mathcal{M} = \langle \mathcal{S}, \mathcal{A}, \mathcal{P}, \mathcal{O}, r, \gamma \rangle$, where $\mathcal{S}$ and $\mathcal{A}$ denote the continuous full state and action spaces, respectively, and $\mathcal{O}$ denotes the partial observation space. The transition dynamics are $\mathcal{P}(\mathbf{s}_{t+1}|\mathbf{s}_t, \mathbf{a}_t)$, the reward function is $r(\mathbf{s}_t, \mathbf{a}_t)$, $\gamma \in [0, 1)$ is the discount factor. The objective is to optimize a policy $\pi_\phi(\mathbf{a}_t | \mathbf{s}_t)$ that maximizes the expected discounted return:
\begin{equation}
    \max_{\boldsymbol{\pi}_\phi} J(\mathcal{M}, \pi_\phi) = \mathbb{E} \left[ \sum_{t=0}^{\infty} \gamma^t r(\mathbf{s}_t, \mathbf{a}_t) \right]
\end{equation}
In this work, we train the actor-critic network using Proximal Policy Optimization (PPO)~\cite{ppo_paper}.

\subsubsection{Policy Network}
The policy network maps robot observations to the action space. The input consists of two components: the proprioceptive observation $\mathbf{o}_t$ and a history of elevation maps $\mathbf{e}_{t-h:t}$. The proprioceptive observation $\mathbf{o}_t$ is a vector obtained directly from sensors, defined as:
\begin{equation}
    \mathbf{o}_t = \left[ \boldsymbol{\omega}_t \quad \mathbf{g}_t \quad \mathbf{c}_t \quad \boldsymbol{\theta}_t \quad \dot{\boldsymbol{\theta}}_t \quad \mathbf{a}_{t-1} \right]^T
\end{equation}
This vector comprises the base angular velocity $\boldsymbol{\omega}_t$, the gravity projection in the body frame $\mathbf{g}_t$, the linear velocity command $\mathbf{c}_t$, the joint positions $\boldsymbol{\theta}_t$, the joint velocities $\dot{\boldsymbol{\theta}}_t$, and the previous action $\mathbf{a}_{t-1}$. The perceptive history $\mathbf{e}_{t-h:t}$ corresponds to a robot-centric local elevation map of size $1.4\;\text{m} \times 1.0\;\text{m}$ at a resolution of $0.05\;\text{m}$. During training, the map generated in Isaac Lab is corrupted with simulated scanning noise before being provided to the actor, and $h$ consecutive frames are stacked to form the perceptive history. We set $h = 5$ throughout this work. For the network architecture, the low-dimensional proprioceptive observation $\mathbf{o}_t$ is processed by a Multilayer Perceptron (MLP), while the elevation map $\mathbf{e}_{t-h:t}$ is processed by a Convolutional Neural Network (CNN). The two resulting feature vectors are concatenated and fed into the actor head, which outputs the action $\mathbf{a}_t$.

\subsubsection{Critic Network}
The critic network estimates the value function $\hat{V}_t$ to guide the policy update. Its input likewise consists of two components: a privileged proprioceptive observation $\mathbf{s}_t$ and noise-free historical elevation maps $\mathbf{e}_{t-h:t}^*$. Compared with the policy input $\mathbf{o}_t$, the privileged observation $\mathbf{s}_t$ additionally includes the ground-truth base linear velocity $\mathbf{v}_t$. The critic network uses the same network architecture as the policy network and outputs the estimated value $\hat{V}_t$.

\subsubsection{Action Space}
The action space $\mathcal{A}$ consists of relative target angles for the robot's joint motors. We define $\mathbf{\boldsymbol{\theta}}_\text{def}$ as the joint positions in which the robot stands stably. The policy output $\mathbf{a}_t$ is added to $\mathbf{\boldsymbol{\theta}}_\text{def}$ to form the desired joint positions $\mathbf{\boldsymbol{\theta}}_\text{des} = \mathbf{\boldsymbol{\theta}}_\text{def} + \mathbf{a}_t$, which are then converted to joint torques $\boldsymbol{\tau}$ through a PD controller:
\begin{equation}
\boldsymbol{\tau} = \mathbf{k}_\text{p} (\mathbf{\boldsymbol{\theta}}_\text{des} - \mathbf{\boldsymbol{\theta}}) + \mathbf{k}_\text{d} (\dot{\mathbf{\boldsymbol{\theta}}}_\text{des} - \dot{\mathbf{\boldsymbol{\theta}}})
\end{equation}
where $\mathbf{k}_\text{p}$ and $\mathbf{k}_\text{d}$ are the proportional and derivative gains. The desired joint velocity $\dot{\mathbf{\boldsymbol{\theta}}}_\text{des}$ is set to zero.

\subsubsection{Reward Function}
The reward function is designed to promote robust locomotion while minimizing unsafe behavior. Table~\ref{tab:rewards} lists all reward terms used in training, where $\mathbf{c}_{i}$ indicates that the $i$-th foot has made contact with the terrain. The total reward is a weighted sum of the individual terms.

\begin{table}[h]
\caption{Reward function terms}
\label{tab:rewards}
\begin{center}
\begin{tabular}{lll}
\toprule
\textbf{Name} & \textbf{Expression} & \textbf{Weight} \\
\midrule
Linear Vel. Tracking & $\exp(-4\|\mathbf{v}_{xy} - \mathbf{v}_{xy}^\text{cmd}\|^2)$ & $5.0$ \\
Angular Vel. Tracking & $\exp(-4\|\boldsymbol{\omega}_z - \boldsymbol{\omega}_z^\text{cmd}\|^2)$ & $5.0$ \\
Base Height & $\exp(-400\|h - h_\text{tar}\|^2)$ & $1.0$ \\
Feet Air Time & $\mathds{1}_{\mathbf{c}_{i}=1} \cdot \min(t_{\text{air}}, t_{\text{c}}, 1.0)$ & $7.0$ \\
Base Lin. Vel. Z & $-\|\mathbf{v}_z\|^2$ & $-1.0$ \\
Base Ang. Vel. XY & $-\|\boldsymbol{\omega}_{xy}\|^2$ & $-0.05$ \\
Base Orientation & $-\|\mathbf{g_\text{xy}}\|^2$ & $-6.0$ \\
Joint Action Rate & $-\|\dot{\mathbf{a}}_{t}\|^2$ & $-0.01$ \\
Joint Action Smooth & $-\|\ddot{\mathbf{a}}_{t}\|^2$ & $-0.01$ \\
Joint Acceleration & $-\|\ddot{\mathbf{\theta}}\|^2$ & $-1e^{-6}$ \\
Joint Torques & $-\|\boldsymbol{\tau}\|^2$ & $-0.2$ \\
Feet Slide & $\sum_{i \in \text{feet}} \mathds{1}_{\mathbf{c}_{i}=1} \|\mathbf{v}_{i, xy}\|^2$ & $-0.1$ \\
Joint Deviation & $-\sum_{j} \|\mathbf{\theta}_j - \mathbf{\theta}_{\text{nom}, j}\|^2$ & $-0.5$ \\
Termination & $\mathds{1}(\text{terminate})$ & $-250.0$ \\
Unsafe stepping & \text{Eq. (10)} & $1.0$ \\
\bottomrule
\end{tabular}
\end{center}
\end{table}

The last entry in the table, labeled Unsafe stepping, corresponds to our proposed unsafe stepping penalty function. When a humanoid robot traverses stairs, two primary types of hazardous foot placement arise: collision between the foot and the stair riser, and landing on the step edge. Existing safety constraints are typically based on physical collision detection, producing sparse reward signals that activate only after the foot has already collided or fully landed, leading to poor training outcomes. To provide continuous guidance before a hazardous step occurs, we design a dense unsafe stepping penalty. This penalty is equally applicable to other terrain types such as slopes and ditches, enabling the policy to be trained with a unified reward function across all terrain configurations.

\begin{figure}[!t]
\centering
\includegraphics[width=0.95\columnwidth]{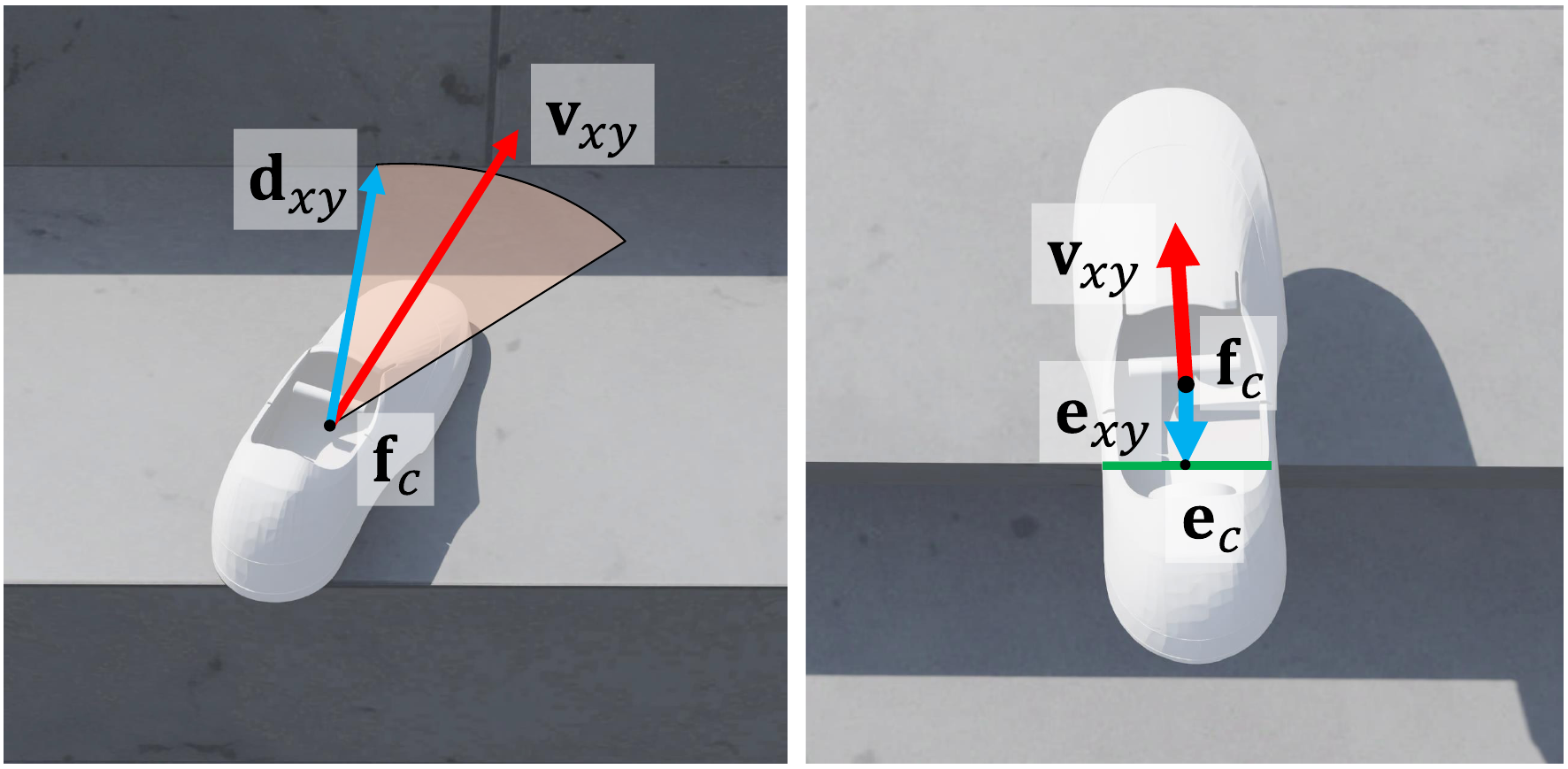}
\caption{
    Illustration of the proposed dense unsafe stepping penalty. The left panel depicts the foot-collision penalty computation, and the right panel depicts the edge-stepping penalty computation.
}
\label{fig:reward}
\end{figure}

For the foot-collision term (left side of Fig.~\ref{fig:reward}), the penalty magnitude increases as the foot approaches an obstacle at higher speed. We first obtain the foot velocity vector ${\mathbf{v}}_{xy}$ (red arrow) in the XY plane, then find the nearest obstacle vector ${\mathbf{d}}_{xy}$ (blue arrow) within a $30^\circ$ cone centered on ${\mathbf{v}}_{xy}$. The collision penalty base value is computed as:
\begin{equation}
p_\text{colli} = \max\left(0, \frac{{\mathbf{v}}_{xy}\cdot{\mathbf{d}}_{xy}}{\|{\mathbf{d}}_{xy}\|}\right)
\end{equation}
When the foot is far from a step, no penalty should be applied. We therefore introduce a safety distance term:
\begin{equation}
d_\text{colli} = \max\left(0, 1-\frac{\|{\mathbf{d}}_{xy}\|}{d_\text{unsafe}}\right)
\end{equation}
where $d_\text{unsafe}$ is the distance threshold below which the penalty takes effect. To distinguish stairs from ordinary slopes, we compute the terrain slope $s$ at the endpoint of ${\mathbf{d}}_{xy}$. When $s$ falls below a threshold $\epsilon_\text{slope}$, the surface is classified as a slope and no penalty is applied. The total foot-collision penalty $r_\text{colli}$ is defined as:
\begin{equation}
r_\text{colli} = -\mathds{1}(s > \epsilon_\text{slope}) \cdot p_\text{colli} \cdot d_\text{colli}
\end{equation}

For the edge-stepping term (right side of Fig.~\ref{fig:reward}), the penalty magnitude similarly increases as the foot approaches a step edge at higher speed. We first obtain the local elevation map directly beneath the foot and compute the gradient magnitude $\mathbf{G} = \|\nabla \mathbf{H}_\text{fl}\|_2$ using a Sobel operator. Points whose gradient magnitude exceeds a threshold are marked as edge points $\mathbf{e}_i$. We then compute the geometric centroid $\mathbf{e}_\text{c}$ (midpoint of the green line in the figure) of all edge points and construct the vector ${\mathbf{e}}_{xy}$ (blue arrow) from the foot center $\mathbf{f}_\text{c}$ to the edge centroid $\mathbf{e}_\text{c}$. Taking the current linear velocity command ${\mathbf{v}}_\text{xy}^\text{cmd}$ (red arrow), the edge penalty $p_\text{edge}$ is:
\begin{equation}
p_\text{edge} = \min\left(0,-\text{sgn}({\mathbf{g}} \cdot {\mathbf{v}}_\text{xy}^\text{cmd}) \cdot {\mathbf{e}}_{xy} \cdot \frac{{\mathbf{v}}_\text{xy}^\text{cmd}}{\|{\mathbf{v}}_\text{xy}^\text{cmd}\|}\right)
\end{equation}
Under this formulation, when the robot ascends stairs, $p_\text{edge}$ takes a negative value and imposes a penalty if the edge lies under the front half of the foot, whereas the penalty becomes zero if the edge lies under the rear half. Because the sign of the projection ${\mathbf{e}}_{xy} \cdot {\mathbf{v}}_\text{xy}^\text{cmd}$ reverses depending on the traversal direction, we introduce the sign correction factor $s_f = -\text{sgn}({\mathbf{g}} \cdot {\mathbf{v}}_\text{xy}^\text{cmd})$, where ${\mathbf{g}}$ is the local mean gradient vector. We further introduce a height weighting term:
\begin{equation}
d_\text{edge} = \max(0, 1 - \frac{\mathbf{d}_{z}}{d_\text{min}})
\end{equation}
where $\mathbf{d}_z$ is the vertical distance from the foot to the ground and $d_\text{min}$ is the minimum height threshold below which the penalty becomes active. The final edge-stepping penalty $r_\text{edge}$ is:
\begin{equation}
r_\text{edge} = p_\text{edge} \cdot d_\text{edge}
\end{equation}
If no edge points are detected in the local elevation map, $r_\text{edge}$ is set to zero.

When the stair tread is narrow, the robot may need to step on the edge to avoid riser collision. We therefore define the final unsafe stepping penalty $r_\text{safe}$ as a weighted sum of the two terms:
\begin{equation}
r_\text{safe} = w_1 \cdot r_\text{colli} + w_2 \cdot r_\text{edge}
\end{equation}
where $w_1$ and $w_2$ are the respective weighting coefficients.

\subsubsection{Terrain Curriculum}

To enable the policy to progressively advance from simple tasks to complex terrains, we employ a terrain curriculum during training. When the robot successfully completes the task at a given difficulty level, it is promoted to the next level; otherwise, it is demoted. We define 10 difficulty levels in total.

The terrain types include stairs, slopes, and flat ground. Stair terrain is divided into ascending and descending variants, with tread depths ranging from $[0.25, 0.6]$\;m and step heights from $[0, 0.23]$\;m. Slope terrain includes ascending and descending configurations with slope angles from $[0, 0.4]$\;rad. The difficulty parameters of all sub-terrains increase linearly with the curriculum level.

%%%%%%%%%%%%%%%%%%%%%%%%%%%%%%%%%%%%%%%%%%%%%%%%%%% 替换方法部分的感知小节

\subsection{Edge-Guided Elevation Map Reconstruction}

To obtain high-frequency elevation maps that preserve precise geometric features, we propose an edge-guided reconstruction framework that operates on sparse point cloud inputs.

\subsubsection{Spatiotemporal Rolling Mapping and Sparse Rasterization}
\label{subsec:mapping_and_async}

\begin{figure}[!t]
\centering
\includegraphics[width=1.0\columnwidth]{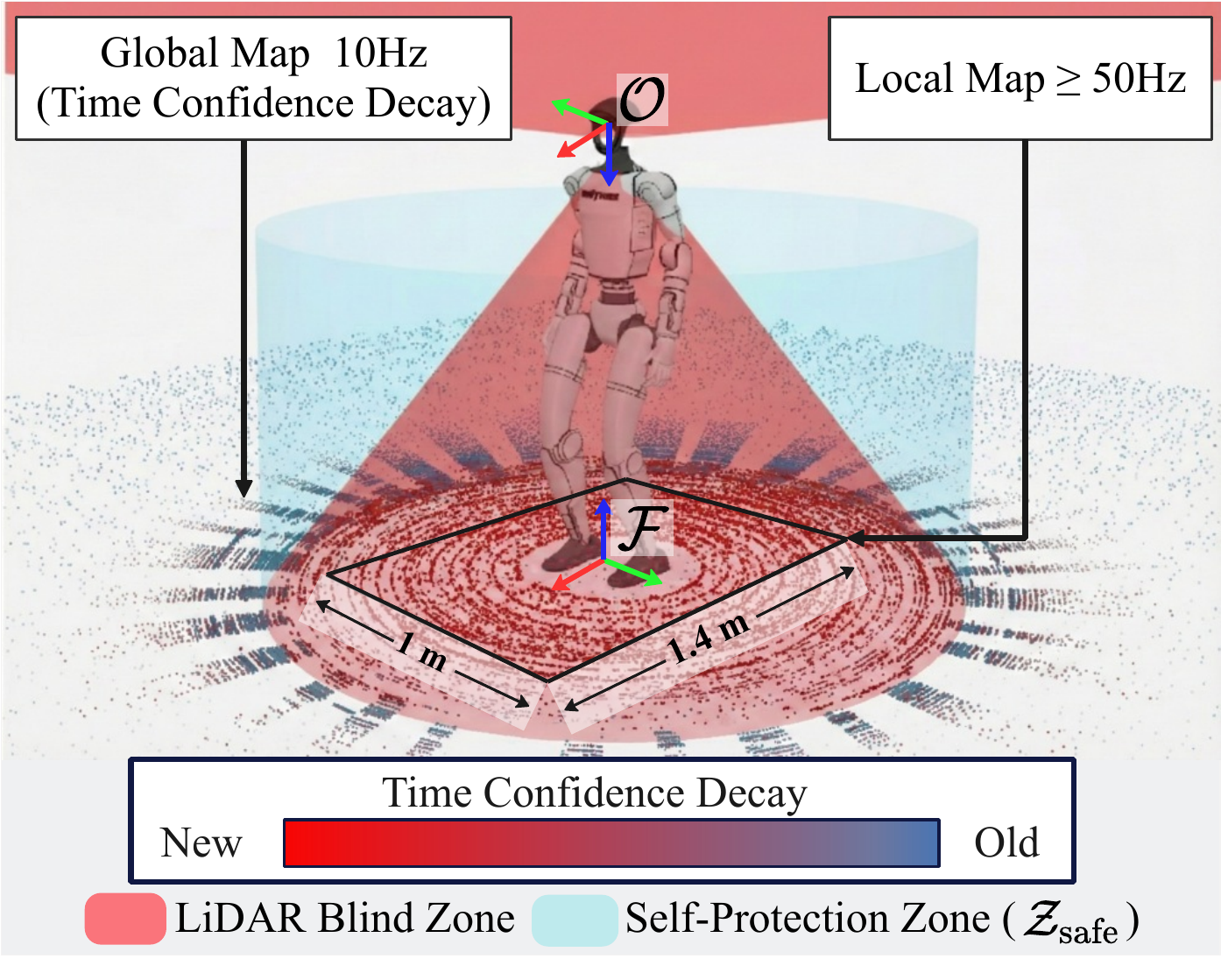}
\caption{
Illustration of the spatiotemporal rolling mapping and the protection zone mechanism. The colors of the point cloud indicate temporal confidence levels. The red conical region represents the physical blind zone of the LiDAR, and the light blue cylinder denotes the self-protection zone $\mathcal{Z}_{\text{safe}}$ constructed beneath the base of the robot.
}
\label{fig:elevation_map}
\end{figure}

We maintain the rolling point-cloud map in the odometry frame $\mathcal{O}$ and filter out returns from the robot's own body. Outside the egocentric protection zone $\mathcal{Z}_{\text{safe}}$ beneath the base, point confidence decays over time. Because timeout-based decay~\cite{eth_elevation} can erase terrain memory in the LiDAR blind zone during slow or in-place motion, points inside $\mathcal{Z}_{\text{safe}}$ retain maximum confidence until leaving it, after which normal decay resumes (Fig.~\ref{fig:elevation_map}).

New LiDAR measurements are fused into the rolling map at 10\;Hz, while accumulated points are reprojected, cropped, and rasterized at 50\;Hz using high-rate odometry. They are transformed into a gravity-aligned, yaw-only local frame $\mathcal{F}$ centered at the ground projection of the base center, with its $x$- and $y$-axes aligned with the heading and lateral directions, respectively, and its $z$-axis aligned with the gravity axis. A $1.4\;\text{m}$ longitudinal $\times\,1.0\;\text{m}$ lateral region is rasterized at $0.05\;\text{m}$ resolution into a $28 \times 20$ map; heights are relative to the local ground, with cellwise maxima retained. The five historical maps are reprojected into the current $\mathcal{F}$ before stacking.

\subsubsection{Edge-Guided Asymmetric Reconstruction Network}
\label{subsec:network}

EGAU takes a two-channel tensor comprising the sparse elevation map and its validity mask. A shared encoder feeds edge and height decoders; the architecture is asymmetric because edge features enter the height decoder at every resolution, with no reverse transfer. This unidirectional guidance preserves boundaries while suppressing cross-edge smoothing. The edge and height heads use sigmoid and linear outputs, respectively.

For decoder level $i \in \{1,2\}$, the height feature is
\begin{equation}
    \mathbf{F}_{\text{h}}^{(i)} = \mathcal{D}_{\text{h}}^{(i)}\!\left(\left[\mathcal{U}_{\text{h}}^{(i)}\!\left(\mathbf{F}_{\text{h}}^{(i-1)}\right),\mathbf{F}_{\text{enc}}^{(i)},\mathbf{F}_{\text{edge}}^{(i)}\right]\right),
\end{equation}
where $\mathbf{F}_{\text{h}}^{(0)}$ is the shared encoder bottleneck, $\mathcal{U}_{\text{h}}^{(i)}$ is learned transposed-convolution upsampling, and $\mathbf{F}_{\text{enc}}^{(i)}$ and $\mathbf{F}_{\text{edge}}^{(i)}$ are the encoder skip and same-level edge features.

\subsubsection{Region-Decoupled Hybrid Loss}
\label{subsec:loss_and_data}

Under sparse LiDAR observations, a globally uniform regression objective is insufficient for accurately recovering terrain structures. In particular, it tends to smooth sharp step boundaries while introducing local height fluctuations in flat regions. We therefore employ a region-decoupled hybrid objective:
\begin{equation}
    \mathcal{L}_{\text{total}} = \mathcal{L}_{\text{h}} + \lambda_{\text{e}} \mathcal{L}_{\text{e}} + \lambda_{\text{r}} \mathcal{L}_{\text{r}} + \lambda_{\text{s}} \mathcal{L}_{\text{s}} + \lambda_{\text{g}} \mathcal{L}_{\text{g}}
\end{equation}
where $\mathcal{L}_{\text{h}}$ and $\mathcal{L}_{\text{e}}$ denote the global height-regression and edge-classification losses adopted from the baseline formulation in~\cite{song2025gait}.

We first compute the gradient magnitude $\mathbf{M}_{\text{gt}} = \|\nabla \mathbf{H}_{\text{gt}}\|_2$ of the ground-truth elevation map using a Sobel operator. Regions with large gradient magnitudes are identified as terrain boundaries, while regions with sufficiently small gradients are treated as locally flat areas, yielding the edge mask $\mathbf{M}_{\text{edge}}$ and flat-region mask $\mathbf{M}_{\text{flat}}$, respectively. These masks are used only during training to impose region-specific reconstruction constraints.

The edge-aware regression loss $\mathcal{L}_{\text{r}}$ applies an additional L1 height-regression penalty within $\mathbf{M}_{\text{edge}}$, reducing height collapse and boundary displacement around stair edges. The smoothness loss $\mathcal{L}_{\text{s}}$ operates exclusively within $\mathbf{M}_{\text{flat}}$, where it suppresses high-frequency height fluctuations by penalizing first-order spatial differences using combined L1 and L2 norms. Finally, the adaptive gradient loss is defined as
\begin{equation}
    \mathcal{L}_{\text{g}} = \frac{1}{N} \sum \left( 1 + \alpha \mathbf{M}_{\text{gt}} \right) \odot \left| \mathbf{M}_{\text{pred}} - \mathbf{M}_{\text{gt}} \right|
\end{equation}
where $\mathbf{M}_{\text{pred}}$ denotes the gradient magnitude of the reconstructed elevation map. The gradient-dependent weighting assigns greater importance to geometrically discontinuous regions, thereby encouraging the network to preserve sharp stair boundaries without amplifying noise over flat terrain.

To reduce the perception gap between simulation and real-world deployment, we additionally introduce a geometry-aware sensor corruption model. Because solid-state LiDAR returns are frequently missing on surfaces with steep local gradients, such as stair risers, gradient-conditioned ray dropping is applied to the simulated point clouds. This augmentation exposes the reconstruction network to structured missing observations similar to those encountered during deployment.

%%%%%%%%%%%%%%%%%%%%%%%%%%%%%%%%%%%%%%%%%%%%%%%%%%% 上面是方法 %%%%%%%%%%%%%%%%%%%%%%%%%%%%%%%%%%%%%%%%%%%%%%%%%%%

\section{RESULTS}

We conduct simulation and real-robot experiments to validate the proposed method. In simulation, we independently evaluate the reinforcement learning locomotion policy and the elevation map reconstruction module. In the real-world experiments, the trained policy is directly deployed on a physical robot that uses the elevation map module for terrain perception in both indoor and outdoor settings.

\subsection{Experimental Setup}

The locomotion policy is trained in Isaac Lab using a single NVIDIA RTX 4090 GPU. The robotic platform used in this paper is the Unitree G1 humanoid robot, which has 29 degrees of freedom. For perception, a Livox Mid-360 LiDAR mounted on the head of the robot provides a $360^\circ \times 59^\circ$ field of view with point clouds. To make the learned walking gait more natural, we additionally employ Adversarial Motion Priors (AMP)~\cite{amp}, whose reference human gait data are collected with a NOKOV optical motion capture system. During real-world experiments, both the RL controller and the elevation map module run on the onboard NVIDIA Jetson Orin NX edge computing platform. Odometry is provided by DLIO~\cite{dlio_paper} at 100\;Hz. The RL controller operates at 50\;Hz.

To comprehensively evaluate the framework, the simulation assessment proceeds along two lines:  ablation and comparison experiments for the locomotion policy and the elevation map reconstruction module. For the locomotion policy comparison, we design four experimental groups:
\begin{itemize}
    \item \textbf{Naive}: Uses the default humanoid training framework provided by Isaac Lab. This baseline employs a simple MLP and does not incorporate the unsafe stepping penalty.
    \item \textbf{Ours w/o penalty}: Adopts the network architecture proposed in this paper but excludes the unsafe stepping penalty.
    \item \textbf{PIM}~\cite{pim_paper}: A single-stage perceptive locomotion policy for humanoid robots that uses elevation maps as perceptive input and trains the network with an hybrid internal model.
    \item \textbf{Ours}: The complete proposed method, which augments Ours w/o penalty with the unsafe stepping penalty.
\end{itemize}

The ablation setup and evaluation metrics for the perception reconstruction module are detailed in Section~\ref{subsec:ablation}.

\subsection{Simulation Experiments}

\subsubsection{Locomotion Policy Experiments}

\begin{figure*}[!t]
        \centering
        \includegraphics[width=1.0\textwidth]{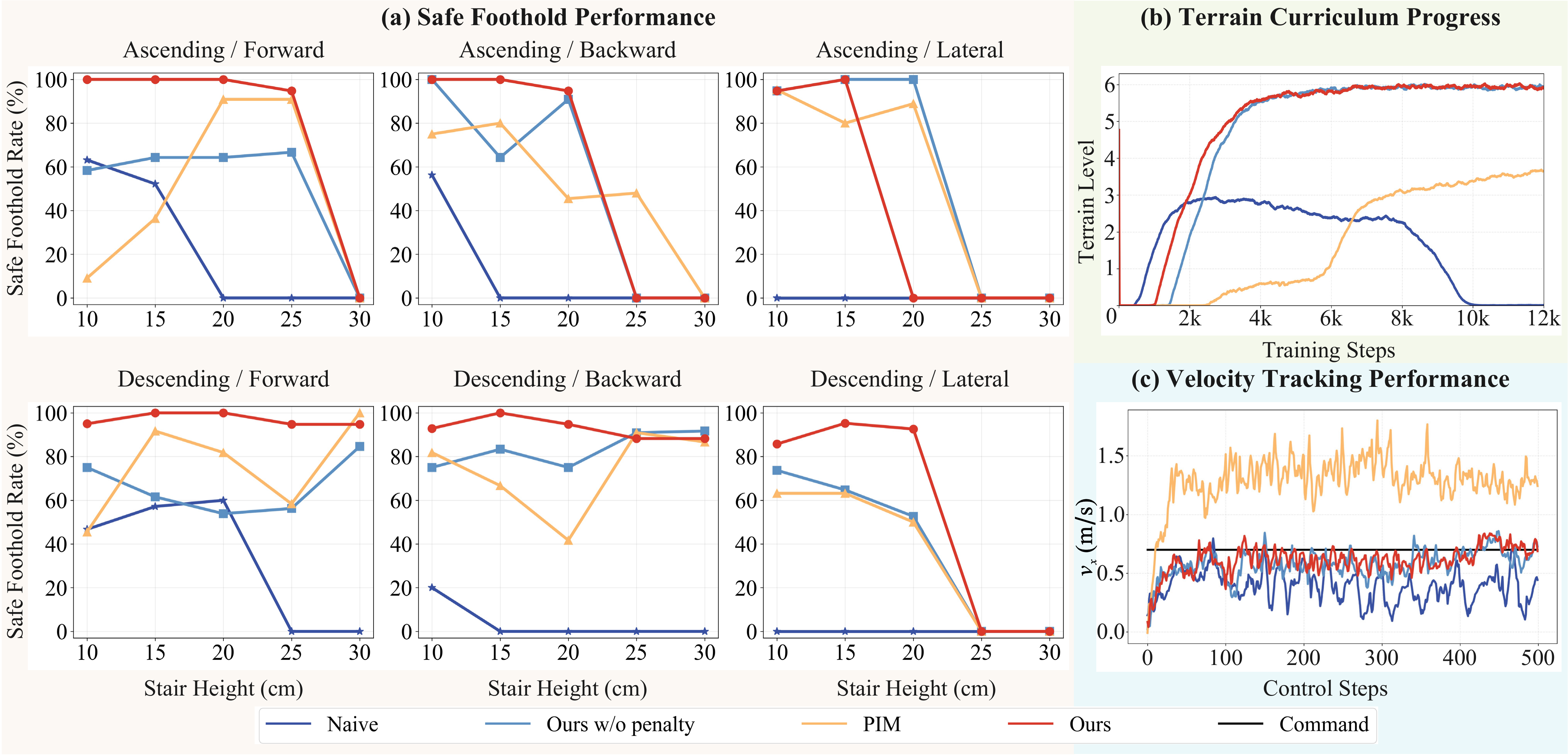}
        \caption{Comprehensive performance comparison in simulation. (a) Safe stepping rate as a function of stair height for each method. The top row corresponds to forward, backward, and lateral stair ascent, while the bottom row shows the corresponding descent results. (b) Terrain curriculum level progression over the course of training. (c) Forward linear velocity tracking when ascending 15\;cm stairs with a commanded velocity of 0.7\;m/s.}
        \label{fig:simulator_curve}
\end{figure*}

We evaluate the overall performance of each method in simulation, including the safe stepping rate, terrain curriculum training curves, and velocity tracking performance, as shown in Fig.~\ref{fig:simulator_curve}.

We examine the safe stepping performance of each method when ascending and descending stairs in different directions, as shown in Fig.~\ref{fig:simulator_curve}(a). The linear velocity command is set to 0.7\;m/s for forward and backward traversal and 0.5\;m/s for lateral traversal. The safe stepping rate is defined as the number of non-hazardous steps divided by the total number of steps. Naive, owing to the simple network structure, fails to learn from the large volume of perceptive information, and the success rate drops as the stair height increases. Both PIM and Ours w/o penalty can traverse stairs exceeding 25\;cm in height; however, without the unsafe stepping penalty, these policies tend to adopt aggressive gaits that disregard foot placement safety, resulting in frequent collisions with stair risers and landings on step edges. Although such aggressive strategies enable the robot to pass extremely challenging stairs in simulation, they severely compromise deployment safety on a physical robot, where a single hazardous footstep can cause hardware damage or a fall.
In contrast, benefiting from the proposed unsafe stepping penalty, Ours learns to lift the feet cautiously and select safe footholds. Although it does not pass certain tests under extreme difficulty settings, it substantially reduces the probability of unsafe foot placements across all passable terrains, significantly outperforming all other methods. Such a low unsafe-stepping probability is crucial for real-world deployment. This confirms that the proposed unsafe stepping penalty effectively guides the policy to traverse stair terrains safely. Notably, in some experiments the safe stepping rate on some lower stairs is lower than that on taller stairs. This is because the policy perceives lower steps as less hazardous and consequently tolerates landing closer to the step edge.

Fig.~\ref{fig:simulator_curve}(b) shows the terrain level progression over 12k training iterations. Naive rises quickly in terrain level during early training owing to the simple architecture, but performance degrades severely as terrain difficulty increases, indicating that a simple MLP alone cannot extract effective terrain features from elevation maps. Ours w/o penalty, which introduces a CNN for elevation map processing, achieves a substantial improvement in performance, reaching approximately level 6 around iteration 4000 and remaining stable thereafter. PIM exhibits considerably slower learning, reaching only about level 3.5 by the end of 12k iterations, likely because the implicit representation learning strategy is inefficient for this complex terrain task. Ours demonstrates the best learning efficiency throughout training, reaching the highest terrain level first at around iteration 3000. This result confirms that the proposed unsafe stepping penalty enables the policy to acquire stair-traversal skills more rapidly.

Finally, Fig.~\ref{fig:simulator_curve}(c) presents the linear velocity tracking performance on 15\;cm stairs in the forward direction. Given a target velocity of 0.7\;m/s, both Ours and Ours w/o penalty track the reference smoothly and accurately. Naive fails to reach the target velocity, and PIM exhibits significant overshoot. This further confirms that the proposed method maintains excellent velocity tracking accuracy while achieving safe foot placement.

\begin{figure*}[!t]
    \centering
    \includegraphics[width=1.0\textwidth]{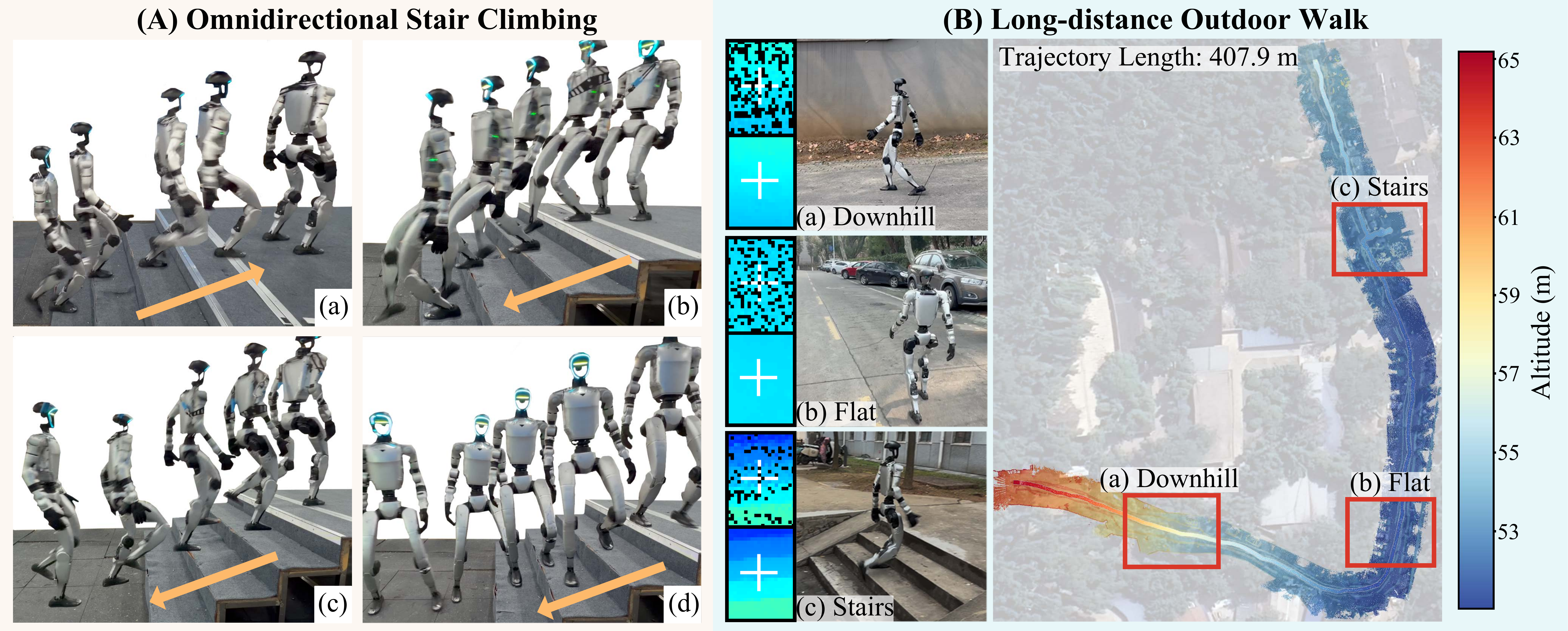}
    \caption{
        Real-world experimental results. (A) The robot traverses stairs in an indoor environment: (a) forward ascent, (b) forward descent, (c) backward descent, and (d) lateral descent. (B) The robot performs long-distance outdoor walking, successfully traversing downhill slopes, flat ground, and stairs. The upper-left inset in each robot image displays the raw elevation map (black dots indicate missing data holes), while the lower-left inset shows the elevation map processed by the proposed reconstruction network.}
    \label{fig:realPic}
\end{figure*}

\subsubsection{Elevation Map Reconstruction Experiments}
\label{subsec:ablation}

To validate the effectiveness of the EGAU architecture and the region-aware loss under conditions with sparse point clouds, ablation experiments are conducted in simulation (see Table~\ref{tab:ablation_net}). All experiments retain the base losses $\mathcal{L}_{h}$ and $\mathcal{L}_{e}$. In addition to the global mean squared error (G-MSE), the edge mask $\mathbf{M}_{edge}$ and the flat-region mask $\mathbf{M}_{flat}$ are utilized to compute the edge mean absolute error (E-MAE) and the flat-region mean absolute error (F-MAE), respectively. Furthermore, the mean spatial gradient of the reconstructed elevation map within flat regions is reported as the flat-region roughness (F-Rgh). This metric provides a comprehensive assessment regarding the capability of the model to preserve step structures and suppress noise within flat areas.

\begin{table}[htbp]
\caption{Ablation results for the reconstruction network and the loss functions}
\label{tab:ablation_net}
\centering
\footnotesize
\setlength{\tabcolsep}{3pt}
\begin{tabular}{c c c c c c c c}
\toprule
\multirow{2}{*}{\textbf{Net}} & \multicolumn{3}{c}{\textbf{Loss}} & \textbf{G-MSE} $\downarrow$ & \textbf{E-MAE} $\downarrow$ & \textbf{F-MAE} $\downarrow$ & \textbf{F-Rgh} $\downarrow$ \\
 & $\mathcal{L}_{r}$ & $\mathcal{L}_{s}$ & $\mathcal{L}_{g}$ & ($10^{-4}\,\mathrm{m}^2$) & ($10^{-2}\,\mathrm{m}$) & ($10^{-3}\,\mathrm{m}$) & ($10^{-2}\,\mathrm{m/cell}$) \\
\midrule
\multicolumn{4}{l}{Baseline} & 2.05 & 1.32 & 5.57 & 2.57 \\
\midrule
EGAU & & & & 1.99 & 1.18 & 5.89 & 2.26 \\
EGAU & \checkmark & & & 1.54 & 0.93 & 5.58 & 2.36 \\
EGAU & \checkmark & \checkmark & & 1.46 & 0.87 & 5.41 & 1.87 \\
EGAU & \checkmark & \checkmark & \checkmark & \textbf{1.23} & \textbf{0.75} & \textbf{4.24} & \textbf{1.32} \\
\bottomrule
\end{tabular}
\end{table}

Table~\ref{tab:ablation_net} shows that EGAU with all region-decoupled losses performs best. For the baseline, we reimplemented the multi-task U-Net of Song et al.~\cite{song2025gait} and retrained it on our LiDAR-derived rasterized dataset using the same data, optimization settings, and evaluation protocol as EGAU. Its higher edge errors indicate over-smoothing under sparse inputs, while EGAU's edge priors reduce E-MAE.

Furthermore, the progressive addition of the region-aware losses targets specific reconstruction artifacts. Specifically, the edge-aware regression loss $\mathcal{L}_{r}$ corrects the absolute height collapse at step edges, while the smoothness loss $\mathcal{L}_{s}$ strictly suppresses high-frequency noise in flat areas, as reflected by a significant reduction in the F-Rgh. Finally, the adaptive gradient loss $\mathcal{L}_{g}$ explicitly penalizes gradient distortions to preserve sharp discontinuities. These results demonstrate that the explicit injection of edge features and the application of region-specific loss constraints fulfill complementary roles in generating accurate and coherent height maps.

\subsection{Real-World Experiments}

The complete perception model runs on the onboard NVIDIA Jetson Orin NX via Onnxruntime-GPU. With only 2.76\;M parameters, the network maintains a stable single-frame inference latency of 2\;ms, easily satisfying the 50\;Hz control loop without competing with the RL controller for resources.

As shown in Figs.~\ref{fig:teaser} and~\ref{fig:realPic}(A), we evaluate the ability of the robot to traverse standard stairs (height 15\;cm, tread width 20\;cm) in different directions. Owing to the simulated ray-drop mechanism, the reconstruction network demonstrates strong robustness to real sensor noise. Guided jointly by the high-quality omnidirectional elevation map and the dense unsafe stepping penalty function, the robot maintains stable gaits across forward, lateral, and backward traversals. By consistently avoiding edge stepping and riser collisions, these results strongly validate the effectiveness of the proposed framework for safe foot placement in real-world deployment.

To verify the performance and stability of the proposed method during long-term motion in outdoor environments, we conducted a continuous, uninterrupted outdoor walking experiment exceeding 400 meters, which was eventually terminated manually, see Fig.~\ref{fig:realPic}(B). Starting from a hillside, the robot traversed a sequence of terrains including a downhill slope, flat ground, and stairs (both ascending and descending). Throughout the entire process, the robot maintained a stable gait and safe foot placement, remaining unaffected by surrounding pedestrians and vehicles. These results demonstrate that the proposed method possesses excellent stability and is capable of long-term operation in complex outdoor environments.

%%%%%%%%%%%%%%%%%%%%%%%%%%%%%%%%%%%%%%%%%%%%%%%%%%% 上面是实验 %%%%%%%%%%%%%%%%%%%%%%%%%%%%%%%%%%%%%%%%%%%%%%%%%%%

\section{CONCLUSIONS}

This paper presents a robust framework for safe omnidirectional humanoid stair traversal. A dense unsafe stepping penalty provides continuous feedback to guide foot placement, significantly enhancing safety and learning efficiency. Additionally, our LiDAR-based mapping with edge-guided reconstruction eliminates blind zones and ensures accurate terrain geometry. Zero-shot sim-to-real experiments on the Unitree G1 confirm reliable omnidirectional mobility on indoor stairs and complex outdoor terrains.

Despite these promising results, this work has certain limitations. The use of omnidirectional 2.5D elevation maps restricts the representation of ditches and similar terrains, and sensor-induced distortions lead to slight performance degradation in real-world deployments compared to simulations.

%%%%%%%%%%%%%%%%%%%%%%%%%%%%%%%%%%%%%%%%%%%%%%%%%%% 上面是结论 %%%%%%%%%%%%%%%%%%%%%%%%%%%%%%%%%%%%%%%%%%%%%%%%%%%

\addtolength{\textheight}{0cm}   % This command serves to balance the column lengths
% \clearpage

\bibliographystyle{IEEEtran}
\bibliography{IEEEabrv,ref}

%%%%%%%%%%%%%%%%%%%%%%%%%%%%%%%%%%%%%%%%%%%%%%%%% 上面是参考文献 %%%%%%%%%%%%%%%%%%%%%%%%%%%%%%%%%%%%%%%%%%%%%%%%%

\end{document}